\documentclass[10pt,twocolumn,letterpaper]{article}

\usepackage{iccv}              % To produce the CAMERA-READY version
\usepackage{multirow}
\definecolor{iccvblue}{rgb}{0.21,0.49,0.74}
\usepackage[pagebackref,breaklinks,colorlinks,allcolors=iccvblue]{hyperref}

\def\paperID{4668} % *** Enter the Paper ID here
\def\confName{ICCV}
\def\confYear{2025}

\title{Adaptive Hyper-Graph Convolution Network for Skeleton-based \\ Human Action Recognition with Virtual Connections}

\author{
Youwei Zhou$^1$, Tianyang Xu$^{1}$\thanks{Corresponding author}, Cong Wu$^1$, Xiaojun Wu$^1$, Josef Kittler$^2$\\
$^1$Jiangnan University, China \ \ \ \ $^2$University of Surrey, UK\\
\tt \small youwei\_zhou@stu.jiangnan.edu.cn\\ 
\tt \small \{tianyang\_xu, congwu, wu\_xiaojun\}@jiangnan.edu.cn\\
\tt \small j.kittler@surrey.ac.uk
}

\begin{document}
\maketitle

\begin{abstract}
The shared topology of human skeletons motivated the recent investigation of graph convolutional network (GCN) solutions for action recognition.
However, most of the existing GCNs rely on the binary connection of two neighboring vertices (joints) formed by an edge (bone), overlooking the potential of constructing multi-vertex convolution structures.
Although some studies have attempted to utilize hyper-graphs to represent the topology, they rely on a fixed construction strategy, which limits their adaptivity in uncovering the intricate latent relationships within the action.
In this paper, we address this oversight and explore the merits of an adaptive hyper-graph convolutional network (Hyper-GCN) to achieve the aggregation of rich semantic information conveyed by skeleton vertices.
In particular, our Hyper-GCN adaptively optimises the hyper-graphs during training, revealing the action-driven multi-vertex relations. 
Besides, virtual connections are often designed to support efficient feature aggregation, implicitly extending the spectrum of dependencies within the skeleton.
By injecting virtual connections into hyper-graphs, the semantic clues of diverse action categories can be highlighted. 
The results of experiments conducted on the NTU-60, NTU-120, and NW-UCLA datasets demonstrate the merits of our Hyper-GCN, compared to the state-of-the-art methods.
% Specifically, we outperform the existing solutions on NTU-120, achieving 90.9\% and 92.0\% in terms of the top-1 recognition accuracy on X-Sub and X-Set.
The code is available at \url{https://github.com/6UOOON9/Hyper-GCN}.
\end{abstract}
\section{Introduction}
Skeleton-based human action recognition is a popular research topic in artificial intelligence, with practical applications in video understanding, video surveillance, human-computer interaction, robot vision, VR and AR~\cite{human_computer_interaction_1, human_computer_interaction_2, VR, skeletonViolenceDetection, wang2023visual, wang2023visual, skeletonVideoSurveillance, skeletonVideoSurvillance2}.
In general, a skeleton sequence contains a series of 2D or 3D coordinates, which can easily be collected by low-cost depth sensors or obtained by video-based pose estimation algorithms \cite{NTU60, NTU120, rajendran2024review, tang2023predicting}. 
Compared to RGB and optical flow images, skeleton data, which represents the basic physical structure of a human being, is of lower dimension, conveying human action with higher efficiency. Moreover, it is robust to illumination changes and scene variations. 
% The information revealed in skeleton data focuses more on human action than on external factors unrelated to the human. 
For these reasons, the adoption of this structural data is very popular in skeleton-based action recognition~\cite{HRNN, DRNN, ST-LSTM, NTU60, InfoGCN, CTR-GCN, DS-GCN}. 
% It is a essentially classification task, where identify which action category the sequence belongs to based on the input coordinate sequence.

\begin{figure}
    \centering
    \includegraphics[width=\linewidth, trim=0mm 25mm 0mm 0mm]{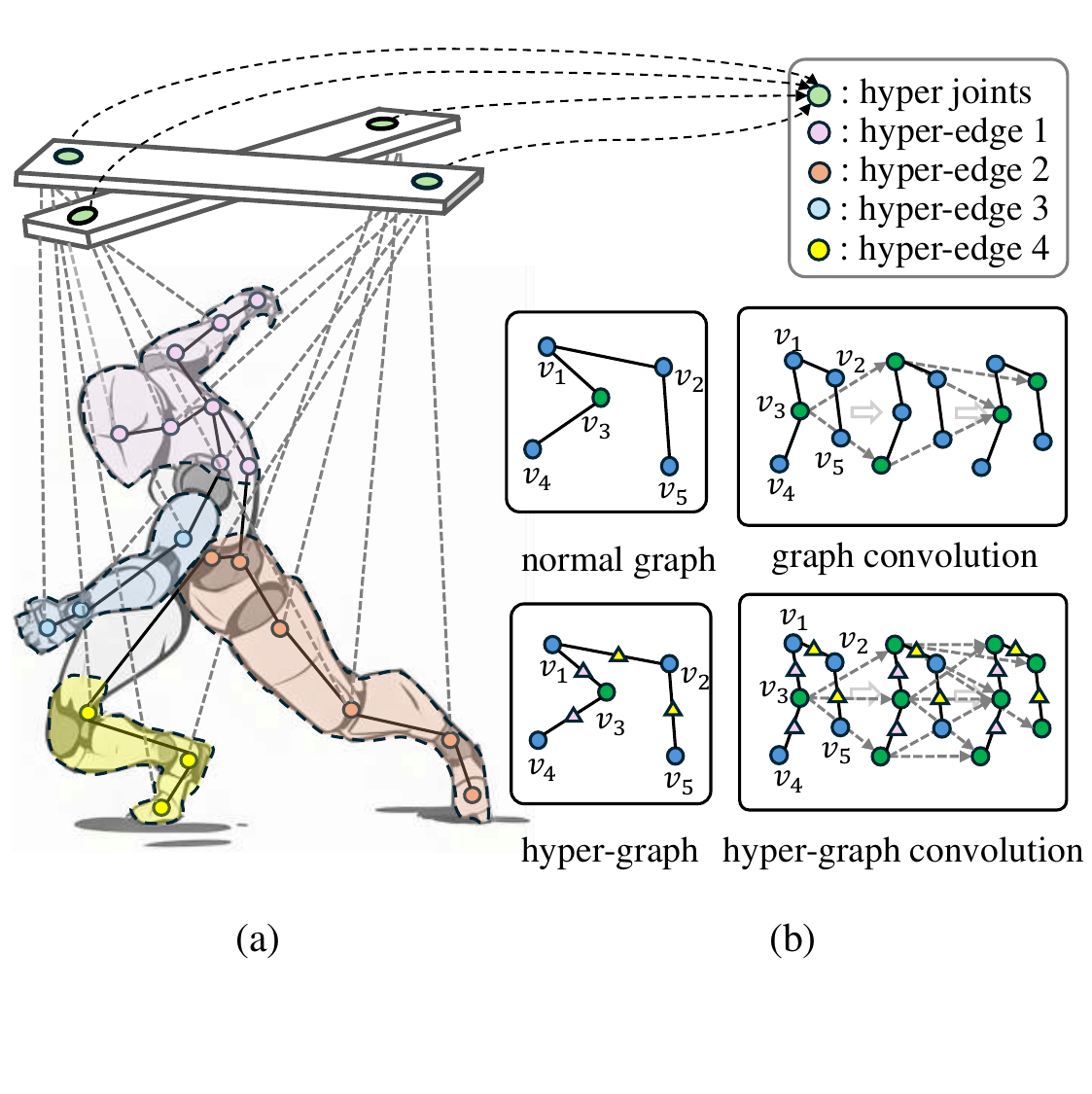}
    \caption{\textbf{Illustration of our Hyper-GCN.} In (a), dotted lines represent the virtual connections. Each coloured part containing multiple joints represents a hyper-graph with hyper-edges. (b) provides an intuitive comparison between normal graph convolution and hyper-graph convolution operations with the same connectivity degree.}
    \label{motivation}
\end{figure}

% \begin{figure*}[ht]
%     \centering
%     \includegraphics[width=\textwidth, trim=5mm 58mm 5mm 55mm]{figures/comparison.pdf}
%     \caption{\textbf{The comparison of normal graph and hyper-graph.} The part (a) is the adjacency matrix of normal graph and the the process of spreading information of graph convolution. The part (b) is the the hyper-graph.} 
    
%     % The vertices contained in same colored hyper-edges represent that these vertices are contained with the same hyper-edge. The part (c) \& (d) is the process of spreading information of the green vertex during the graph convolution and hyper-graph convolution.}
%     % The part (d) is the process of spreading information of the green vertex during the hyper-graph convolution. The pink triangles denote the 2-uniform hyper-graph, which represent each hyper-edge contain only 2 vertices.}
%     \label{comparison}
% \end{figure*}

% In early stage, most of scholars dedicate to extract the information of skeleton sequence by traditional machine learning methods.
To facilitate skeleton-based action recognition, both Recurrent Neural Networks (RNNs)~\cite{HRNN, DRNN, ST-LSTM, NTU60} and Convolutional Neural Networks (CNNs)~\cite{P_CNN, 2s-CNN, 3dCNN} have been well explored.
However, RNNs themselves cannot depict the intrinsic skeleton topology.
The learned filters of CNNs, on the other hand, neglect the spatio-temporal structure of the skeleton.
Drawing on these observations, recent studies have focused on how to model the skeleton topology directly. 
In principle, the physical topology of human joints and bones can be consistently represented by a graph. 
Accordingly, the graph convolutional network (GCN) is typically introduced to aggregate feature information conveyed by skeleton joints~\cite{ST-GCN, HD-GCN, BlockGCN}.
In terms of data representations, skeleton joints are also one-dimensional sequence data, similar to language. 
Therefore, with the rise of recent self-attention techniques~\cite{Transformer, VIT}, several Transformer-based action recognition solutions \cite{ST-TR, TSTE, DSTA-Net, IIP-Transformer} have been released. 
Transformer considers skeletal joints as tokens and uses attention maps to reflect the topology. 
These methods have achieved excellent performance but sacrifice much higher GFLOPS and parameters than GCN.

% In terms of modelling advances, owing to the outstanding feature extraction and function fitting capabilities of deep learning, it has become the mainstream topic in skeleton-based human action recognition. Specifically, Recurrent Neural Networks (RNNs)~\cite{Du_2015_CVPR, Veeriah_2015_ICCV, Shahroudy_2016_CVPR, Liu_2018_TPAMI} are employed for skeleton-based human action recognition, due to their ability to capture the skeleton temporal information through cyclic connections, which is highly effective for understanding the human action. But RNNs fail to exploit adequately the topology inherent in the human skeleton. 

% The human action is constrained by the skeleton, the coordinated functioning of various joints is integral to every movement. Therefore, it is unreasonable to view them solely as discrete nodes. Thus, leveraging the inherent topological relationships of the human skeleton to enhance the model's learning of the distinct semantic information between joints is critical. Due to the importance of topology, GCNs stand out in skeleton-based human action with its structure, that enables extracted feature in joints through the topology of skeleton.

% many attempts are based on GCNs~\cite{ST-GCN, 2s-AGCN, CTR-GCN, InfoGCN, HD-GCN, DS-GCN}. 
% Because GCNs allow information to aggregate between discrete vertices using a given topology such as the human skeleton. 
In general, two neighbouring joints can pass messages through their shared bone.
In graph terms, this corresponds to the exchange of information between two vertices along their connecting edge.
Besides the physical skeleton topology, attempts have been made to explore the implicit relationships among joints~\cite{2s-AGCN, CTR-GCN, InfoGCN, HD-GCN}, suggesting different variants of skeleton topology.
% , aiming to uncover non-physical but interactively significant implicit relationships among joints. 
In principle, these approaches assume a binary connection between each connected vertex pair. 
Mathematically, the constructed topology in the adjacency matrix is represented by a normal graph. 
However, human actions are jointly defined by several joints. 
Hence, human actions encompass not only binary relations between vertex pairs but also multi-joint relationships. 
For instance, the action primitive \textit{starting running} is manifest in the human raising the left hand while the right leg steps forward, as shown in Figure~\ref{motivation} (a). 
The binary connections are not sufficient to capture the synergistic interaction of multiple joints. This strongly argues for constructing feature aggregation paths involving multiple vertices.

% For example, when running, the  body employs both hands and feet, swinging the left hand as the right foot is taken forward, and swinging the right hand as the left foot is taken forward. Some human action encompass not just the topological relationships between pair of joints, but the topology of a group including all the joints. Therefore, merely identifying the adjacency between pairs of joints may not accurately capture the coordinated functioning of multiple relation among joints.

Accordingly, we propose to construct a hyper-graph to depict the skeleton topology and take advantage of the outstanding performance of hyper-graph analysis techniques ~\cite{dynamicHypergraph, hypergraphAttention, multilevelHypergraph, HGNN+}. 
The hyper-graph topology involves multiple node  connections, rather than binary connections of a normal graph. 
As shown in Figure~\ref{motivation} (a), in a hyper-graph, a hyper-edge can link more than two vertices. This linking has the capacity to represent complex collaborative relations among human joints. 
As one hyper-edge associates multiple vertices, a single hyper-graph convolution enables aggregating all the features along the hyper-edges.

An illustration is provided in Figure~\ref{motivation} (b), where the normal graph and hyper-graph are presented to demonstrate their differences in passing the information conveyed by the vertex features.
% Conceptually, for a fair comparison, we typically set each hyper-edge to contain only 2 vertices, which is the same as the normal graph.
% for a fair comparison.) with the normal graph. 
In the normal graph convolution, after 2-layer aggregation, the information of the green vertex is spread to 2 other vertices. 
In contrast, in a hyper-graph convolution, the information of the green vertex spreads to all the vertices, creating an extended receptive field.
% That's almost twice as much in terms of quantity. 
Theoretically, by modelling a skeleton using a hyper-graph for the aggregation of joint information, improved efficiency can be obtained during learning the action semantics. 

% This is very effective for understanding human action. For this, there are further validation in the subsequent experiments.
% In particular, we emphasize that different topology need to be constructed for different human actions, rather than adopting the same topology for each sample. For example, in the action of clapping hands, the interaction between the joints represented by the two palms will better reflect the characteristics of this action than the joints represented by other parts.

% As shown in \textbf{Figure~\ref{comparison} (c)\&(d)}, the hypergraph contains the higher-order topology between joints, allowing for more efficient feature interaction between joints.

% Furthermore, for construct the hyper-graph in a smooth way without compromising the semantic integrity of the original features, Hyper-GCN translates the original features into a low-dimensional subspace for metrics to distinguish the hyper-edge to which each joint belongs. In addition, for the final topology involved in feature aggregation, we retain the original physical topology to combine with the constructed hyper-graph. Because the human actions cannot be accomplished beyond the original physical structure, it is not optimal to treat them as completely arbitrary combinable joints.

However, the key to achieving enhanced information aggregation by hyper-graph convolution lies in the conformity of the hyper-graph structure to the real topology of human action. 
Several studies \cite{HypergraphNN, SHypergraphGCN} attempt to adopt a hyper-graph to model the skeleton topology, mostly based on superficial prior knowledge and an unfounded definition of the structure of hyper-graph. 
% Some studies on normal graphs \cite{2s-AGCN, CTR-GCN, InfoGCN, DS-GCN} have demonstrated that the topology representing human action should be more adaptive to  different human actions. 
Distinguishing our approach from the existing hyper-graph-based methods, we design an adaptive solution for constructing the hyper-graph with virtual connections. 
It can construct a Non-uniform hyper-graph specific to each individual action type.

Besides the information aggregation issue, the capacity of an action recognition system is also constrained by its input features.
In general, the number of skeleton joints is fixed in existing benchmarks, \textit{e.g.}, 25 for NTU-120~\cite{NTU120}.
The entire recognition process relies on a successful capture of the interactions among the skeleton joints.
In the domain of artistic puppetry, the history of driving actions, such as Shadow Play~\footnote{\url{https://en.wikipedia.org/wiki/Shadow_play}} and Marionette~\footnote{\url{https://en.wikipedia.org/wiki/Marionette}} goes back more than 2000 years.
This kind of art form provides an inspiration for involving additional 'hyper joints' which can drive or facilitate a better communication between existing joints.
The underlying spirit is to alleviate the pressure on the real joints to store and transfer complex semantics.
Jointly with the hyper joints, real joints can focus more on storing neighbouring joint features and relegating the task of transferring the global information to the hyper joints. 
Interestingly, the class token in existing Transformers can also be considered as a hyper-token
% which can  information carrier   we notice that several approaches utilize Transformers 
~\cite{Transformer, VIT}.
% for recognition. The class token connects adaptively with other tokens to extract the global information, which is widely applied in the Transformers. Inspired by this, we have renewed our attention to the significance of global information among selected joints for classification. And We believe that the global information introduced should be consistent in its form of representation with the vertex information extracted from physical joints, such as the class token. 
% However, such connections, which are missing structured information, are not suitable for graph convolution. Thus, we introduce hyper joints with virtual connections. 
As shown in Figure~\ref{motivation}, the skeleton is described as a marionette, where the actions are "controlled" by connecting hyper joints to the real joints. 
This suggests that the hyper joints are not only able to capture the representation information of human action, but also reveal the implicit information between physically connected joints as hints for recognition.

% As mentioned above, the addition of hypergraph and hyper joints accelerates the information interaction between joints. As a result, the variance of semantic information increases rapidly between each layer of the model. When the semantic information changes too fast, it will lead to the catastrophic oblivion of the deep features extracted by the model to the shallow ones. Therefore, as shown in \textbf{Figure.~\ref{motivation} (a)}, we mitigate this problem by dividing the backbone into 3 stages based on different channels and introduce dense connection within each stage. In addition, we tend to aggregate the feature represented by the physical joints and the feature represented by the hyper joints to collaboratively discriminate the classification to which each sample belongs to.

By endowing an adaptive non-uniform hyper-graph with hyper joints, virtual connections are created to perform comprehensive hyper-graph convolutions.
We construct Hyper-GCN based on the above design principles.
Extensive experiments on 3 datasets, NTU RGB+D 60~\cite{NTU60}, NTU RGB+D 120~\cite{NTU120}, and NW-UCLA~\cite{NW-UCLA}, are conducted for evaluation. 
The results validate the merits of our proposed Hyper-GCN. 
The main contributions are as follows:

\begin{itemize}
\item An adaptive non-uniform hyper-graph to represent the human skeleton topology.
% and combine it with the original physical topology to construct topology. 
% Compared with the normal graph, the efficiency of feature interaction is improved via multi-vertex aggregation. 
Compared with a fixed hyper-graph, the constructed topology is more action-specific, thus boosting the discrimination.
\item{The injection of virtual hyper joints, enriching the connectivity of the physical joints, from a global semantic perspective.}
% that are beneficial for discriminating the human action.}
% \item{We introduce dense connection in the model to fuse the shallow features and deep features between layers of the backbone.}
\item{As the processing architecture, we propose a hyper-graph convolution network (Hyper-GCN). The SOTA performance achieved on three public datasets demonstrates the merits of the Hyper-GCN and virtual connection designs.}
% In particular, it outperforms the SOTA on NTU-RGB+D 120.}
\end{itemize}

\begin{figure*}[t]
    \centering
    \includegraphics[width=\textwidth, trim= 20mm 50mm 15mm 20mm]{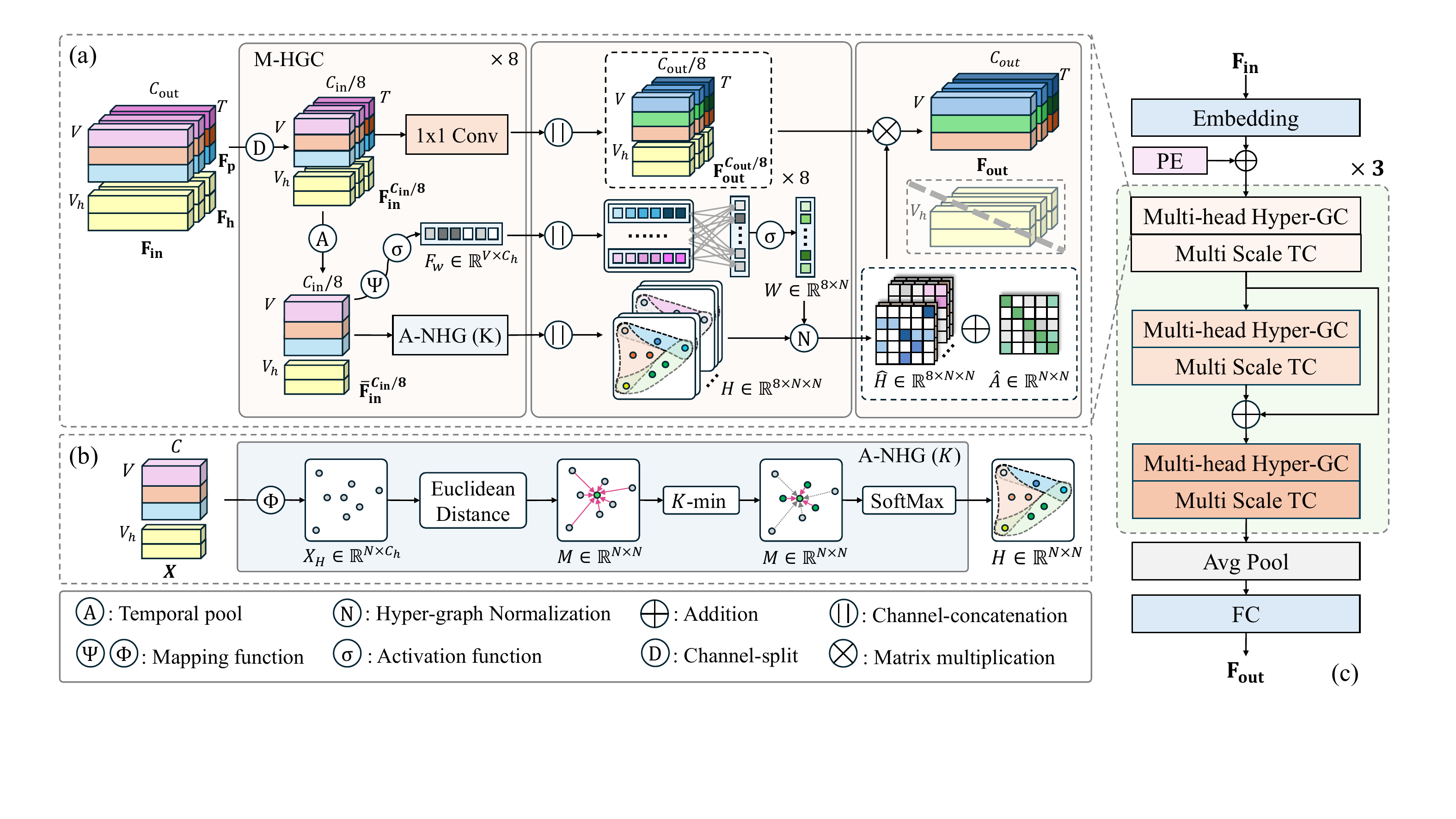}
    \caption{\textbf{The framework of our Hyper-GCN.} Part (a) represents the Multi-head Hyper-graph Convolution (M-HGC) module. Part (b) represents the process of constructing an adaptive hyper-graph. Part (c) represents the architecture of Hyper-GCN. $\mathbf{F_p}$ represents the features of real joints. $\mathbf{F}_{h}$ represents the hyper-joints which are learnable parameters. $\hat{A}$ represents the physical topology.}
    \label{HyperGC}
\end{figure*}

\section{Related Work}
\subsection{Graph Topology for Action Recognition}
% 1. graph setting

A graph can represent a human skeleton, preserving the joint relationships via predefined edges.
To aggregate semantics, GCNs~\cite{SGN, DDGCN, DGNN, DC-GCN+ADG, MST-GCN, AS-GCN, EEGCN, BlockGCN} and Transformers ~\cite{ST-TR, STST, DSTA-Net, TSTE, skateformer} have been well studied.
% in recent years~\cite{SGN, DDGCN, DGNN, DC-GCN+ADG, MST-GCN, AS-GCN, EEGCN, BlockGCN}. The recent Tranformers~\cite{ST-TR, STST, DSTA-Net, TSTE, skateformer} are also outstanding.

For GCN-based approaches, ST-GCN~\cite{ST-GCN} proposes to represent the topology with an adjacency matrix of 3 subsets by modeling the spatio-temporal relevance.
Similarly, 2s-AGCN~\cite{2s-AGCN}, InfoGCN~\cite{InfoGCN}, and DS-GCN~\cite{DS-GCN} use a self-attentive mechanism to learn the topology from joint pairs. 
Besides the intuitive spatio-temporal dimensions, CTR-GCN~\cite{CTR-GCN} proposes to learn the topology for channels to refine the skeleton. 
% HD-GCN~\cite{HD-GCN}, on the other hand, represents the topology manually, decreasing the generalisation power.}

For Transformer-based solutions, IIP-Transformer \cite{IIP-Transformer} adopt a self-attention mechanism to establish the intra-part and the inter-part joints. 
IGFormer \cite{IGformer} learns the topology between persons at both semantic and distance levels. 
STTFormer \cite{STTFormer} proposed structure can capture the correlation between joints in consecutive frames. 

All the above two paradigms construct the topology from joint pairs, which rely only on the binary relation between two vertices.
In their implementation stage, the adjacency matrix or attention map is used to represent the topology. 
In this case, high-order information among joints is not taken into consideration, neglecting the collaborative power among multiple joints. 
Though hyper-graph is considered to construct topology using multivariate joint relationships, current~\cite{SHypergraphGCN, HypergraphNN} solutions manually set the hyper-graphs, which greatly relies on human experience, sacrificing the adaptability of graph learning. 

% This does not lead to a better adaption to various human actions.

\subsection{Feature Configurations for Action Recognition}
% 2. input setting
% GCNs propose various modules to extract features better
It has been observed that the input features play the most essential role in delivering high recognition accuracy~\cite{Shift-GCN, MS-G3D, EfficentGCN-B4, EEGCN, partGCN, IIP-Transformer, IGformer, STTFormer} in GCNs and Transformers. 

Drawing on this, Graph2Net~\cite{Graph2Net} proposes to extract local and global spatial features jointly. 
CTR-GCN~\cite{CTR-GCN} uses multi-scale temporal convolution to extract temporal features. 
While HD-GCN~\cite{HD-GCN} introduce the hierarchical edge convolution to key edge features. 
To extend the perception field, STC-Net~\cite{STC-Net} uses the dilated kernels for graph convolution to capture the features. 

Similarly, IG-Transformer \cite{IGformer} aggregate the features of two persons as one to mine the semantic information.
STST \cite{STST} split the coordinate features, semantic features and temporal features into three kinds of tokens.
TSTE \cite{TSTE} extracting and merging the spatial and temporal features as input to transformer encoders. 

After all, existing methods~\cite{ST-GCN, 2s-AGCN, InfoGCN, DS-GCN, multiGCN, DSTA-Net, ST-TR} receive the input features from real skeleton joints. 
However, each skeleton joint acts as an information carrier during forward passing, which is required to deliver both local context and global semantics.
Given a fixed representative capacity (number of channels), we believe it is necessary to involve additional virtual joints to balance the pressure of storing local and global information.

\section{Approach}
% In this section, we introduce the detailed structure of Hyper-GCN. we first describe the normal graph convolution and hyper-graph convolution. Then, we introduce the Adaptive Hyper-graph Construction Module (AHC-Module). Next, we introduce the hyper joints with virtual connections. Finally, we introduce the model architecture.

\subsection{Preliminaries}
In general, the input features in graph convolution are represented by $\mathbf{{F}_{in}}\in\mathbb{R}^{C\times T\times V}$, where $C$ represents the number of channels in the feature maps. 
Given the topology of a human skeleton, we usually define the graph $\mathcal{G}=(\mathcal{V},\mathcal{E})$, where $\mathcal{V}$ represents the set of joints and $\mathcal{E}$ represents the set of edges between joints. 
For the set of edges $\mathcal{E}$, it is formulated as an adjacent matrix $A \in\mathbb{R}^{N\times N}$, where $N$ represents the number of human joints.
The normalised adjacency matrix is represented by $\hat{A} \in\mathbb{R}^{N\times N}$. 
The normalisation operation is formulated as follows: 
\begin{equation}
    \hat{A}=\Lambda^{-\frac{1}{2}}\ A \Lambda^{-\frac{1}{2}},
\end{equation}
where $\Lambda$ represents the diagonal matrix stored with the degree of every joint. 
% This operation constrains the aggregation of features between joints by row normalization and column normalization of the adjacency matrix $A \in\mathbb{R}^{N\times N}$. In addition, ST-GCN~\cite{ST-GCN} propose that split the spatial graph to 3 subsets which are represented by $S=\{s_{id}, s_{cf}, s_{cp}\}$. $s_{id}$, $s_{cf}$ and $s_{cp}$ denote the identity, centrifugal, and centripetal joint subsets. Thus the normal graph convolution leverages 3 adjacent matrices to aggregate the features. 
The entire normal graph convolution process can be formulated as:
\begin{equation}
\mathbf{F_{out}}=\sigma(\hat{A} \mathbf{F_{in}}P),
    \label{eq-2}
\end{equation}
where $P \in \mathbb{R}^{C \times C'}$ is the learnable parameters, representing the feature transformation patterns in the feature space. 
$\sigma$ denotes the non-linear activation function ReLU.

% As shown in \textbf{Figure.~\ref{comparison} (a)}, a normal graph connects only 2 different human joints per edge, which represents that an edge can only describe the interrelationship between two joints. However, the action of human body is often generated by the interactions between various joints. As shown in \textbf{Figure.~\ref{comparison} (b)}, the major difference between a hypergraph and graph is that an edge of a hypergraph called hyperedge can contain various joints instead of only 2.

\subsection{Hyper-graph}
\label{sec: hyper-graph}
Here, we use $\mathcal{G_H}=(\mathcal{V_H},\mathcal{E_H}, \mathcal{W_H})$ to define the spatial hyper-graph~\cite{HGNN+} with human skeleton. 
$\mathcal{V_H}$ and $\mathcal{E_H}$ follow similar definitions in the normal graph, which represent the set of joints and the set of hyper-edges.
In addition, we introduce $\mathcal{W_H}$ to represent the weights of each hyper-edge. 
Since a hyper-edge contains multiple joints, the corresponding topology can no longer be simply represented by an adjacency matrix. Therefore, we introduce the incidence matrix $H \in \mathbb{R}^{N \times M}$ to describe the topology of each joint in the hyper-graph. 
$N$ represents the number of joints and $M$ represents the number of hyper-edges. 
Given $v \in \mathcal{V_H}$ and $e \in \mathcal{E_H}$, the values of incidence matrix be determined by:
\begin{equation}
h(e, v) =
    \begin{cases}
        1, & v \in e,\\
        0, & v \notin e
    \end{cases}.
\end{equation}
% As shown in \textbf{Figure.~\ref{comparison} (b)}, for example, if the $H_{ij}$ equals 1, which means that the $i$-th joint is contained in the $j$-th hyper-edge. The reverse is true.

Similar to the normal graph convolution, it also needs to normalise the hyper-graph to modulate the aggregated features. 
The degree of hyper-graph consists of the degree of joints and the degree of hyper-edges. 
The degree of joints is represented by the sum of weights of all joints contained in each hyper-edge. 
Given $v\in \mathcal{V_H}$, it can be described as follows:
\begin{equation}
    d(v)\ = \sum_{e\in\mathcal{E_H}}{w(e)h(v,e)},
\end{equation}
where we use a diagonal matrix $D_v \in \mathbb{R}^{N \times N}$ to represent the degree of joints. 
$W\in\mathbb{R}^{M \times M}$ represents the weight matrix of hyper-edges, which is formulated as a diagonal matrix. 
The degree of hyper-edges represents the sum of the number of joints contained in each hyper-edge. 
Given $e \in \mathcal{E_H}$, it can be described as follows:
\begin{equation}
    d(e)\ = \sum_{v\in\mathcal{V_H}}{h(v,e)},
\end{equation}
where we use a diagonal matrix $D_e \in \mathbb{R}^{M \times M}$ to represent the degree of hyper-edges. 
Therefore, the normalisation of a hyper-graph is formulated as follows:
\begin{equation}
    \hat{H} = D_v^{-1} H W D_e^{-1} H^T,
    \label{eq-6}
\end{equation}
where $\hat{H} \in \mathbb{R}^{N \times N}$ represents the normalised incidence matrix for hyper-graph convolution. 
% The complete hyper-graph convolution is similar with Eqn.~\eqref{eq-2}, which is formulated as:
% \begin{equation}
%     \mathbf{F_{out}} = \sigma(\hat{H} \mathbf{F_{in}}W).
% \end{equation}

% It can also be formulated with row normalization and column normalization as:
% \begin{equation}
%     \hat{H} = D_v^{-\frac{1}{2}} H W D_e^{-1} H^T D_v^{-\frac{1}{2}},
% \end{equation}
% where it needs to make sure that the value in weight matrix $W\in\mathbb{R}^{M \times M}$ can not be negative. The shape of the matrix is consistent with that of a normal graph and does not require additional storage. So the normalized incidence matrix also can be mulitplied directly to the input feature without additional format conversion. So the integrity operation of the hypergraph convolution is described as: 
% \begin{equation}
%     \mathbf{F_{out}}=\sigma(\sum_{s\in S}{D_{s,v}^{-1} H_s W_s D_{s,e}^{-1} H_s^T \mathbf{F_{in}}W_s}).
% \end{equation}

\begin{figure}[t]
    \centering
    \includegraphics[width=\linewidth, trim=0mm 65mm 20mm 30mm]{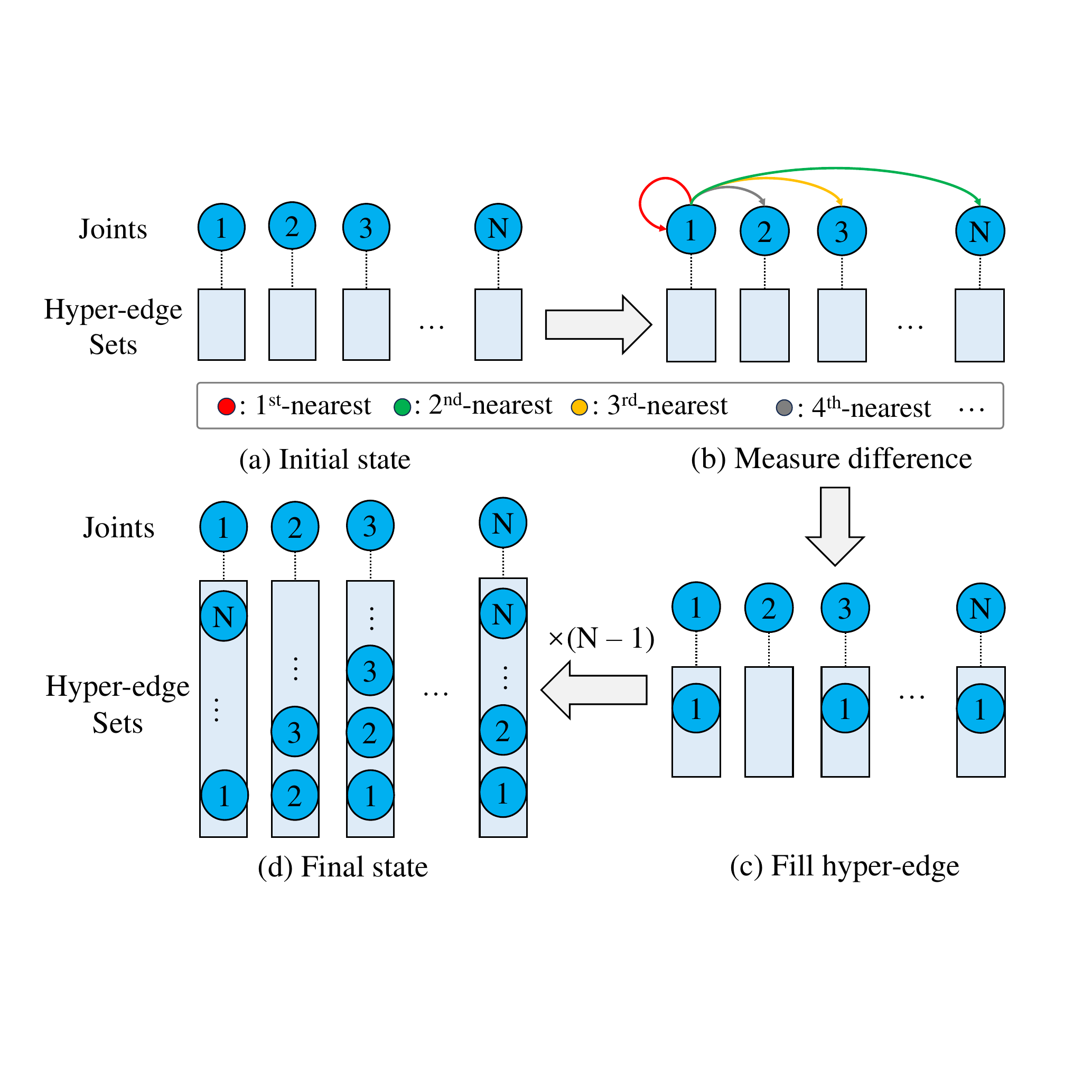}
    \caption{\textbf{Instance of the incidence matrix construction by A-NHG with hyper-parameter $K=3$.} $N$ represents the number of joints. (a) represents the hyper-edge sets are null. (b) represents that measure the distance between 1-st joint and other joints. (c) represents that retrain $K$-nearest hyper-edges which contain 1-st joint. (d) represents the final state.}
    \label{A-NHG}
\end{figure}

\subsection{Adaptive Non-uniform Hyper-graph}
Constructing the hyper-graph incidence matrix is critical to understanding human actions.
% On the one hand, single utilization of the original physical topology is not enough, which is not able to uncover the implicit collaborative interaction between non-adjacent joints. On the other hand, constructing the hyper-graph in an artificially given and empirical way is too limited to cope with a wide variety of human actions. Therefore, we expect that our model can learn how to construct the most appropriate hyper-graph based on different input features.
As shown in Figure~\ref{HyperGC} (b), we design an Adaptive Non-uniform Hyper-graph (A-NHG)  Construction. 

Given the general features $\mathbf{X}\in\mathbb{R}^{N \times C}$ of spatial skeleton joints, $N$ represent the number of joints, $C$ represent the channels. Our A-NHG construct the hyper-edge with each joint as the center of mass, and a total of $N$ hyper-edges are constructed. A-NHG utilises the Euclidean distance to measure the joint difference. 

% These embedded joints represent vertices as well as hyper-edges. 
% For the mapping of hyper-edges and vertices, the reason we apply only one mapping function instead of two separate ones is to ensure that vertices and hyper-edges are within the same semantic space. 
We use one mapping function $\Phi \in \mathbb{R}^{C \times C_h}$ to embed the features into the subspace for constructing the hyper-graph. $C_h$ represents the hidden channels of the mapping subspace. This design enables preserving the original features to reflect their spatial relevance. 
Based on $X_H \in \mathbb{R}^{N \times C_h}$, we define the set of joints $\mathcal{V_H} = \{v_1 \in \mathbb{R}^{C_h}, v_2 \in \mathbb{R}^{C_h}, ... , v_N \in \mathbb{R}^{C_h}\}$. Given $v_i, v_j \in \mathcal{V_H}$, each element $m_{i,j}$ in the distance matrix $M\in\mathbb{R}^{N \times N}$ can be obtained as follows:
\begin{equation}
    m_{i,j} = m_{j,i} = ||v_i - v_j||_2.
\end{equation}
% where $N$ represent the number of embedding joints, and the $C$ represent the number of channel in embedding joints.

% \begin{figure}
%     \centering
%     \includegraphics[width=\linewidth, trim= 15mm 60mm 15mm 45mm]{figures/virtual-connection.pdf}
%     \caption{\textbf{The subsets of hyper joints and their virtual connections.} The red joints denote the central joints in the skeleton. The green joints denote the hyper joints. The blue joints denote the physical joints connected with hyper joints.}
%     \label{virtual-connection}
% \end{figure}

In principle, we need to transform the distance matrix into a probability incidence matrix, which determines whether the specific joints are contained with the same hyper-edges. 
In order to guarantee that the model is trainable, we use the softmax operation to assign the probabilities. 
However, directly using the softmax would inevitably result in each joint belonging to all hyper-edges to some extent. 

Since this outcome is undesirable, we introduce constraints on each joint to limit the number of hyper-edges it can belong to, preventing excessive connections. 
For one joint, we retain only the $K$ nearest hyper-edges represented by joints in the semantic space for probability estimation. And $K$ is the hyper-parameter of A-NHG. 

Given the row vector $\mathbf{m}_i$ in the distance matrix $M$, which represents the distances between joint $i$ and the other joints. We select the indices of the minimum $K$ joints from $\mathbf{m}_i$ and add them to the set $set_{i}$.
% \rr{Given the set of joints $d_{i,j}$ in distance matrix $Dis \in \mathbb{R}^{N \times N}$, 
Based on $set_{i}$, $h_{i,j}$ in the incidence matrix $H \in \mathbb{R}^{N \times N}$, can be calculated as follows:

\begin{align}
    h_{i,j} & = 
    \begin{cases}
        \frac{\text{exp}(-m_{i,j})} {\sum_{k \in set_{i}}{\text{exp}(-m_{i,k})}}, & j \in set_{i}, \\
        \ \ \ \ \ \ \ \ \ \ \ \ \ \ 0, & j \notin set_{i}.
    \end{cases}
\end{align}

% The shape of our incidence matrix is consistent with the adjacency matrix of normal graph. While the topology with hyper-graph contains more high-order information.
In contrast to uniform hyper-graph (every hyper-edge contains the same number of joints), our approach limits the maximum number of every joint that can be contained by hyper-edges. 
Obviously, the number of joints contained in each hyper-edge constructed in this way is non-fixed, allowing the hyper-edges to capture more differentiated associations.
The entire process for example with $K$ = 3 is illustrated in the Figure~\ref{A-NHG}.

\subsection{Multi-head Hyper-graph Convolution}
\label{sec: M-HGC}
% As our A-NHG can depict the topology into a hyper-graph by adjusting the hyper-parameter of $K$. 
To accommodate the semantic information reflected by different group channels, we further propose the Multi-head Hyper-graph Convolution (M-HGC), as shown in Figure~\ref{HyperGC} (a). 
 
We use separate branches to independently perform hyper-graph convolution on the topologies represented by the multi-head hyper-graphs, thereby enhancing the computational efficiency of the Hyper-GCN. For instance, the features $\mathbf{F_{in}}\in\mathbb{R}^{C_\text{in} \times T\times V}$ are divided into $8$ separate branches along the channel dimension, which are processed by $8$ M-HGC parallelly, delivering $8$ separate hyper-graphs.
% while minimising the loss of information from the temporal dimension. 

Next, we perform temporal pooling, and then the obtained spatial information is the evidence to determine the optimal hyper-graph construction. 
This can efficiently decouple the temporal and spatial clues. 
After performing temporal average-pooling, we can obtain $\mathbf{\bar{F}_{in}}\in\mathbb{R}^{C_\text{in} \times V}$. 

% To adjust the scale of the input feature channels, we adopt a layer normalisation operation on the channel dimension. 
The hyper-graph is depicted by the incidence matrix and weight matrix. 
So, we use a separate mapping function for each head to embed the features into the subspace for constructing the hyper-graph. 
This design enables preserving the original features to reflect their spatial relevance.
After that, we introduce the A-NHG to obtain the incidence matrix. For the weight matrix, we adopt MLP to measure each hyper-edge.
% In addition, we incorporated an activation function for the weight of each hyper-edge, allowing it to be constrained within a certain range.}
The hyper-graph obtained operation can be formulated as follows:
\begin{align}
    H & = \uplus^8_{k=1}\text{A-NHG}(\mathbf{\bar{F}^\mathnormal{k}_{in}}), \\
    W & = \sigma_2(\Psi_2(\uplus^8_{k=1} \sigma_1(\Psi^k_1(\mathbf{\bar{F}^\mathnormal{k}_{in}})))),
\end{align}
where $\Psi_1 \in \mathbb{R}^{C_\text{in}/8 \times C_h}, \Psi_2 \in \mathbb{R}^{(8 \times C_h) \times 8}$ are the mapping functions, which are set as learnable parameters.
$C_h$ represents the hidden channels of the mapping subspace.
$k$ represents the channels $[(k-1)\times C_\text{in}/8+1, ..., k\times C_\text{in}/8]$ of $\mathbf{\bar{F}_{in}}$.
$\uplus$ represents the channel concatenation. 
% and $W_\textrm{emb}$ represents the weights of hyper-edges.
% $\textrm{Pooling}_\textrm{T}$ denotes performing pooling in the temporal dimension.
% LN denotes the layer normalisation along the channel dimension.
% $\Phi\in\mathbb{R}^{C/8 \times C_h}$ is the mapping function, which is set as learnable parameter. $C_h$ represents the channels of the mapping subspace. 
$\sigma_1, \sigma_2$ denote the activation function LeakeyReLU and Tanh to obtain the weight of each hyper-edge, limiting the values with the range of $[-1, 1]$. Based on $H \in \mathbb{R} ^{8 \times N \times N}, W \in \mathbb{R}^{8 \times N}$, normalise the hyper-graph by sec.\ref{sec: hyper-graph} to obtain the $\hat{H} \in \mathbb{R}^{8 \times N \times N}$.

Besides the hyper-graph, we incorporate the physical topology to emphasise the natural physical relations of a human being's skeleton. 
To achieve this objective, existing methods~\cite{ST-GCN, 2s-AGCN, CTR-GCN, HD-GCN} divide the physical topology of the human body into 3 subsets. 
They are represented by $S=\{s_{id}, s_{cf}, s_{cp}\}$, where $s_{id}$, $s_{cf}$ and $s_{cp}$ denote the identity, centrifugal, and centripetal joint subsets. 
In our M-HGC, to ensure that the integrated physical topology in each head remains complete, we merge them into a single set.
% \rr{To ensure that the integrated physical topology in each branch remains complete, we abandon the approach of dividing the physical topology into 3 subsets $s_{id}$, $s_{cf}$ and $s_{cp}$ denote the identity, centrifugal, and centripetal joint subsets.}. 
% \rr{Based on $K$, we divide MS-HGC into 8 separate branches, each corresponding to a different scale of hyper-graph convolution.}
8 heads are aggregated by channel concatenation after the hyper-graph convolutions, as shown in Figure~\ref{HyperGC} (a). 
The operation of M-HGC can be formulated as follows:
\begin{align}
    % \mathbf{F^{'}_{in}} & = \mathbf{F^{[(i-1) \times C^{'} + 1, ..., i \times C^{'}]}_{in}}, \\  
    \mathbf{F_{out}} & = \uplus^8_{k=1} {(\hat{A_k} + \alpha \cdot \hat{H_k}) \mathbf{F^\mathnormal{k}_{in}} P_k},
\end{align}
where $\hat{A}$ and $\hat{H}$ denote the normalized physical adjacency matrix and normalized incidence matrix. 
$P$ is the learnable weight parameter for feature transform.
$\alpha$ is a learnable parameter for the topology fusion of each head.
% The following normalisation operation is defined in Eqn.~\eqref{eq-6}. 
% the hyper-graph normalization is formulated as follows:
% \begin{equation}
%     \hat{H} = D_v^{-1} H W_{embedding} D_e^{-1} H^T,
% \end{equation}
% where the $D_v^{-1}$ represents the diagonal matrix stored the degree of joints and the $D_e^{-1}$ represents the diagonal matrix stored the degree of hyper-edges. 
% Besides, since human actions are inherently limited by the skeleton, in order to mitigate the topology forgetting, we fuse the constructed hyper-graph with the original physical topology.

\subsection{Virtual Connections}
It is worth emphasising that incorporating learnable joints among different samples is essential for enhancing the model capacity, as these learnable joints capture generalised features of human actions. 
This not only enriches the semantic information but also facilitates easier interaction connections among real joints. 
Therefore, we introduce the hyper-joints which are to participate in the hyper-graph convolution as shown in Figure~\ref{HyperGC} (a). 

The shape of hyper-joints is consistent with the features of physical joints $\mathbf{F_{p}} \in \mathbb{R}^{C \times T\times V}$, which is described as $\mathbf{F_{h}} \in \mathbb{R}^{C \times T\times V_h}$ ($V_h$ represents the number of hyper-joints). 
To ensure alignment in the temporal dimension, the hyper-joints are shared across each frame. Specifically, the learning of hyper-joints is supervised by the loss function, which guides these hyper-joints to support generalisable driven features embedded within large amounts of data.
Furthermore, we set independent hyper-joints at each layer of Hyper-GCN, aiming to harmonise features at different depths. 
Typically, these hyper-joints are involved in spatial hyper-graph convolution rather than in temporal convolution.

To diversify these hyper-joints, 
% we emphasise the need for a certain level of diversity among the hyper-joints learned from the semantic information. To address this, 
we propose the Divergence Loss for hyper-joints optimisation to mitigate their homogenisation. We adopt a cosine matrix $C \in \mathbb{R}^{V_h \times V_h}$ to measure the differences between hyper-joints, which can be formulated as follows:
\begin{equation}
    C = \frac{\mathbf{F_{h} F^\mathnormal{T}_{h}}}{||\mathbf{F_{h}}||^2},
    \label{Eq-14}
\end{equation}

% offset the difference of hyper-joints, as it is always 1 and cannot be optimised.}
% The underlying reason of this design 
% This design is because, under the supervision of classification signals, hyper-joints are difficult to learn predictive features for varied human action. 
Specifically, in Divergence Loss, we measure the differences of hyper-joints with the mean of the cosine matrix $C$ in each layer. Since the correlation of hyper-joints themselves cannot be optimised, we manually eliminate this part by minus the $V_h$.
The loss calculation can be formulated as:
\begin{align}
    \mathcal{L}_h(C) & = \frac{\sum^{V_h}_{i=1}\sum^{V_h}_{j=1} ReLU(c_{i,j}) - V_h}{V_h(V_h - 1)}, \quad c_{i,j} \in C, \\
    \mathcal{L} & = \mathcal{L}_{CE} + \frac{1}{L}\sum^L_{l=1} L_h(C_l),
\end{align}
where $V_h$ represents the number of hyper-joints we introduced. 
$C_{l}$ represents the cosine matrix of the $l$-th layer. $\mathcal{L}_\text{CE}$ represents the cross-entropy loss.
% Where $\mathbf{I}$ represents the identity matrix which is used to emphasise the irrelevance between each learned hyper-joint pair. 
% Additionally, to preserve the original physical topology, we use the A-NHG solely to learn the topology between feature $\mathbf{F_h}$ and feature $\mathbf{F_{p}}$.
Additionally, we manually connect the hyper joints to all the physical joints, which optimises the topology from the datasets.
% The weights of the connections are learnable by the network during training. 
% For the topology of a normal graph $\mathcal{G} = (\mathcal{V}, \mathcal{E})$. 
% We create several hyper joints with different connection preferences.
% The first hyper joint 
% We expand the skeleton from the central joint to the neighbouring joints,\xu{ and at each step of expansion, the joints reached, we identify as a subset.} 
% As shown in Figure~\ref{virtual-connection}, beginning with the second step, we partition them into $k$ subsets for each hyper joint. 
% Given subset $k$, we add joint $p^k$ into vertices set and $\{(p^k, j^k_1), ..., (p^k, j^k_2), ..., (p^k, j^k_{N_k})\}$ into edges set $\mathcal{E}$, where $k$ represents the number of subsets being partitioned. 
% $N_k$ represents the number of the physical joints which are connected with the hyper joint in subset $k$. 
% For hyper-graph, the topology with hyper joints is constructed by AHC-Module.

\subsection{Entire Architecture}
Hyper-GCN consists of an embedding layer~\cite{InfoGCN}, 9 spatial-temporal convolution layers as shown in Figure~\ref{HyperGC} (c). Each layer consists of the proposed M-HGC and Multi Scale Temporal convolution (MS-TC)~\cite{CTR-GCN}.
9 Layers are categorised into 3 stages. In each stage, we introduce the dense connections to integrate deep and shallow features in each stage. 
% which can smooth the obtained distribution against the joints interaction at each layer.
Additionally, to further validate the potential of Hyper-GCN, we propose a base version and a large version. The channels in each stage are set to 128, 256, 256 for the base version and 128, 256, 512 for the large version. 
% \noindent \textbf{The embedding layer} is utilised to map individual joints into a high-dimensional feature space for subsequent adaptive hyper-graph convolution. 
% % However, the embedding layer leads to a loss of the positional topology corresponding to the original joints.
% To preserve the position awareness of the embedded features, we introduce the position embedding in the embedding layer inspired by InfoGCN~\cite{InfoGCN}.
% % In addition, because the embedding layer leads to a loss of the positional topology corresponding to the original joints. Therefore, we embedding layer and position embedding are also involved in Hyper-GCN inspired by InfoGCN~\cite{InfoGCN}.  

% \noindent \textbf{The 9 layers of Hyper-GCN} are categorised into 3 stages. 
% Each stage consists of 3 layers. 
% Each layer consists of Multi-Scale Hyper-graph convolution (M-HGC) and Multi-Scale Temporal convolution (MS-TC)~\cite{CTR-GCN}. 
% In stage 1, the number of channels is set as 128. 
% Then it increases to 256 for stage 2 and 512 for stage 3. 

% \noindent \textbf{Dense connections} are introduced to integrate deep and shallow features in each stage, which can smooth the obtained distribution against the joints interaction at each layer.

\begin{table*}[ht]
\centering
\scalebox{0.76}{
\begin{tabular}{ccccccccccc}
\hline
\multirow{2}{*}{Category} & \multirow{2}{*}{Methods} & \multirow{2}{*}{Modalities} & \multirow{2}{*}{Params (M)} & \multirow{2}{*}{GFLOPs} & \multicolumn{2}{c}{NTU-RGB+D 60} & \multicolumn{2}{c}{NTU-RGB+D 120} & \multirow{2}{*}{NW-UCLA (\%)} \\

& & & & & X-Sub (\%) & X-View (\%) & X-Sub (\%) & X-Set (\%) & \\ \hline

\multirow{8}{*}{GCN}
& ST-GCN~\cite{ST-GCN} & J+B & - & - & 81.5 & 88.3 & 70.7 & 73.2 & - \\
& 2s-AGCN~\cite{2s-AGCN} & J+B & - & - & 88.5 & 95.1 & 82.5 & 84.2 & - \\
% & DGNN~\cite{DGNN} & CVPR19 & J+B+JM+BM & 89.9 & 96.1 & - & - & - \\
% & SGN~\cite{SGN} & CVPR20 & J+B+JM+BM & 89.0 & 94.5 & 79.2 & 81.5 & 92.5 \\
% Shift-GCN~\cite{Shift-GCN} & CVPR 2020 & J+B+JM+BM & & & 90.7 & 96.5 & 85.9 & 87.6 & 94.6 \\
& DC-GCN+ADG~\cite{DC-GCN+ADG} & J+B+JM+BM & 4.9 & 1.83 & 90.8 & 96.6 & 86.5 & 88.1 & 95.3 \\
% & DDGCN~\cite{DDGCN} & ECCV 2020 & J+B+JM+BM & 91.1 & 97.1 & - & - & - \\
& MS-G3D~\cite{MS-G3D} & J+B+JM+BM & 2.8 & 5.22 & 91.5 & 96.2 & 86.9 & 88.4 & - \\
& MST-GCN~\cite{MST-GCN} & J+B+JM+BM & 12.0 & - & 91.5 & 96.6 & 87.5 & 88.8 & - \\
& CTR-GCN~\cite{CTR-GCN} & J+B+JM+BM & 1.5 & 1.97 & 92.4 & 96.4 & 88.9 & 90.6 & 96.5 \\
& EfficientGCN-B4~\cite{EfficentGCN-B4}  & J+B+JM+BM & 2.0 & 15.20 & 91.7 & 95.7 & 88.3 & 89.1 & - \\
& InfoGCN~\cite{InfoGCN} & J+B+JM+BM & 1.6 & 1.84 & 92.7 & 96.9 & 89.4 & 90.7 & 96.6 \\
& FR Head~\cite{FR-Head} & J+B+JM+BM & 2.0 & - & 92.8 & 96.8 & 89.5 & 90.9 & 96.8 \\
& HD-GCN*~\cite{HD-GCN} & J+B+J'+B' & 1.7 & 1.77 & 93.0 & 97.0 & 89.8 & 91.2 & 96.9 \\
& DS-GCN~\cite{DS-GCN} & J+B+JM+BM & - & - & 93.1 & \underline{97.5} & 89.2 & 90.3 & - \\
& BlockGCN~\cite{BlockGCN} & J+B+JM+BM & 1.3 & \textbf{1.63} & 93.1 & 97.0 & 90.3 & 91.5 & 96.9 \\ \hline

\multirow{3}{*}{Transformer}
& DSTA-Net~\cite{DSTA-Net} & J+B+JM+BM & 3.5 & 16.18 & 89.5 & 95.7  & 86.6 & 89.0 & - \\
& IIP-Transformer~\cite{IIP-Transformer} & J+B+JM+BM & 2.9 & 7.20 & 89.5 & 95.7 & 89.9 & 90.9 & - \\
% & HyperFormer~\cite{} & - & J+B+JM+BM & - & - & - & - & - \\
& SkateFormer~\cite{skateformer} & J+B+JM+BM & 2.0 & 3.62 & \underline{93.5} & \textbf{97.8} & 89.8 & 91.4 & \textbf{98.3} \\
\hline

\multirow{5}{*}{HGCN}
& Hyper-GNN~\cite{HypergraphNN} & J+B+JM+BM & - & - & 89.5 & 95.7 & - & - & - \\
% & DHGCN~\cite{} & - & J+B+JM+BM & 90.7 & 96.0 & 86.0 & 87.9 & - \\
& Selective-HCN~\cite{SHypergraphGCN} & J+B+JM+BM & - & - & 90.8 & 96.6 & - & - & - \\
& DST-HCN~\cite{DST-HCN} & J+B+JM+BM & 3.5 & 2.93 & 92.3 & 96.8 & 88.8 & 90.7 & 96.6 \\ 
& Ours (B) & J+B+JM+BM & \textbf{1.1} & \textbf{1.63} & 93.3 & 97.4 & \underline{90.5} & \underline{91.7} & 97.2 \\

% \\[-5pt]

% \hline
& Ours (L) & J+B+JM+BM & 2.3 & 2.88 & \textbf{93.7} & \textbf{97.8} & \textbf{90.9} & \textbf{92.0} & \underline{97.6} \\ \hline
\end{tabular}
}
\caption{\textbf{Comparison of Hyper-GCN with advanced solutions on NTU-RGB+D 60, NTU-RGB+D 120, and NW-UCLA datasets.} For a fair comparison, we use a 4-streams ensemble to evaluate the performance of the mentioned methods. J, B, JB, and JM represent the joint, bone, joint motion, and bone motion. In addition, for HD-GCN~\cite{HD-GCN}, we use the results from 4-streams according to the original papers for a fair comparison. The \textbf{bold} font represents the best result and \underline{underline} font represents the 2-nd best result.}
\label{coparison_sota}
\end{table*}

\section{Evaluation}
% In this section, we conduct extensive experiments on standard benchmarking datasets to evaluate our Hyper-GCN, in the task of skeleton-based human action recognition. 
% We compare Hyper-GCN with existing approaches and find that Hyper-GCN is at a level close to state-of-the-art. 
% Further, we performed ablation experiments on the Adaptive Hypergraph Construction Module (AHC-Module), the Hyper-graph Convolution (Hyper-GC), the inclusion of hyper joints, and the backbone as a way to verify the effectiveness of our designs. 
% In addition, For the AHC-Module, we explored its construction patterns. For the inclusion of hyper joints, we verify the performance of the model with the number of hyper joints introduced. In the 2 existing problems mentioned, we also analyze whether Hyper-GCN can be further solved.
\subsection{Datasets}
\noindent \textbf{NTU-RGB+D 60 \& 120} NTU-RGB+D 60~\cite{NTU60} is a large dataset widely used in skeleton-based human action recognition, which is categorised into 60 classes. NTU-RGB+D 120~\cite{NTU120} is an extention to 120 classes of NTU 60. 4 benchmarks recommended by the official are adopted: (1) NTU60-XSub, (2) NTU60-XView, (3) NTU120-XSub, (4) NTU120-Xset.
% It consists of 56,880 different samples, categorised into 60 classes, obtained from the directed performances of 40 different actors. 
% There are two 2 evaluation benchmarks. (1) Cross-Subject (X-Sub): For the 40 actors, 20 are used for training and 20 for validation. (2) Cross-view (X-View): For 3 views, 2 are used for training, and 1 for validation.

% \noindent \textbf{NTU-RGB+D 120} NTU-RGB+D 120~\cite{NTU120} is an extention of NTU 60, which introduce 57,367 new action samples. 
% It consists of 114,480 different samples, categorised into 120 classes instead of 60 classes. 
% The number of actors increase to 106. 
% It also corresponds to 2 benchmarks. (1) Cross-Subject (X-Sub): For the 106 actors, 53 are used for training and 53 for validation. (2) Cross-Setup (X-Set): For the 32 setups, the sample with even setup IDs are used for training, and the odd setup IDs for validation.

\noindent \textbf{Northwestern-UCLA.} The Northwestern-UCLA (NW-UCLA) dataset~\cite{NW-UCLA} contains 1494 video clips, which is categorised into 10 classes. It contains 3 different camera views and is performed by 10 actors. 
% As the proposed benchmark, for 3 camera views, 2 camera views are used for training, and the remaining 1 is used for validation.

\subsection{Implementation Details}
% We have conducted experiments on 3 public datasets NTU-RGB+D 60, NTU-RGB+D 120 and NW-UCLA~\cite{NTU60, NTU120, NW-UCLA}.
Our training and evaluation stages are on a single GPU RTX 3090.
In the training phase, Hyper-GCN is optimised by Stochastic Gradient Descent (SGD) with Nesterov momentum set at 0.9 and a weight decay at 0.0004. 
Our implementation uses label smooth cross-entropy loss with the Divergence Loss we proposed. We set a total of 140 epochs with the start 5 warm-up epochs. The initial learning rate is 0.05, which is reduced to 0.005 at epoch 110 and to 0.0005 at epoch 120. 
% The training batch size is set to 64 for NTU RGB+D and NTU RGB+D 120, and 16 for NW-UCLA. 
% Our training strategy is based on CTR-GCN, InfoGCN~\cite{CTR-GCN, InfoGCN}.

\subsection{Comparison with the State-of-the-Art}
Multi-stream ensemble proposed in ~\cite{Multi-Streams} have been proven effective by most of the existing state-of-the-art methods.
We also use a 4-stream ensemble to evaluate the base version and the large version of Hyper-GCN. The detailed results are reported in Table \ref{coparison_sota}.
% 4-streams ensemble contains the joint, bone, joint motion, bone motion. Joint is the most general form of data used in skeleton-based human action recognition, where each point of data input to the model represents the coordinates of a 3D joints. Bone is the transformation of joints from coordinates to vectors through the differentiation of adjacent joints, which further represents the length of a bone in 3-dimensional space. Joint motion represents the difference of joints in the time dimension in the Joint modality, which can represents the motion of joints. Similarly, the Bone motion represents the motion in the Bone modality.

In summary, the base version of Hyper-GCN comprehensively outperforms all GCN-based and HGCN-based (hyper-graph utilised) SOTA and surpasses the Transformer-based SOTA on the NTU120, with a most lightweight design. 
Moreover, when scaled to match the parameters of the lightest Transformer-based method, the large version of Hyper-GCN achieves 1-st place in 4 benchmarks and 2-nd place in 1 benchmark.
% \xu{We report the training results of the 4-streams in for form of weighted summarion to get the final results.} 
% To the best of our knowledge, a comparison with SOTA shows that our proposed Hyper-GCN achieves the SOTA level on each dataset. 
% Especially, we outperform the involved competitors on NTU-RGB+D 120. 
 % In addition, Hyper-GCN has reached a high level in the case of only using joint and bone modalities for ensemble. 
This demonstrates the effectiveness and merits of Hyper-GCN.

% \begin{table}
% \centering
% \scalebox{0.86}{
% \begin{tabular}{cccccccccc}
% \hline
% \multicolumn{3}{c}{Branch-1} & \multicolumn{3}{c}{Branch-2} & \multicolumn{3}{c}{Branch-3} & NTU-RGB+D 120 \\
% N        & F       & P       & N        & F       & P       & N        & F       & P       & X-Sub (\%)    \\ \hline
%          &         & \checkmark       &          &         & \checkmark      &          &         & \checkmark       & 85.1          \\
% \checkmark        &         &         & \checkmark        &         &         & \checkmark        &         &         & 86.1 ($\uparrow$ 1.1)         \\
%          & \checkmark       &         &          & \checkmark       &         &          & \checkmark       &         & 86.1 ($\uparrow$ 1.1)          \\
%          & \checkmark       &         &          & \checkmark       &         &          &         & \checkmark       & 86.3 ($\uparrow$ 1.2)          \\
% \checkmark        &         &         & \checkmark        &         &         &          &         & \checkmark       & 86.2 ($\uparrow$ 1.1)          \\
% \checkmark        &         &         &          & \checkmark       &         &          &         & \checkmark       & \textbf{86.5 ($\uparrow$ 1.4)}         \\ \hline
% \end{tabular}
% }
% \caption{\textbf{Comparison of the adaptive hyper-graph construction patterns.} Branch-1, Branch-2, Branch-3 denote the three parallel hyper-graph convolutions of Hyper-GC. N, F and P denote NJC, FJC and physical topology.}
% \label{comparison_AHC}
% \end{table}

\subsection{Ablation Study}
We conduct ablation experiments with visualisations to analyse the effectiveness of our design. \textbf{red}{All the ablation study is evaluated with the joint modality of base version.}

\begin{table}
\centering
\begin{tabular}{cccc}
\hline
& & \multicolumn{2}{c}{NTU RGB+D 120 X-Sub (\%)} \\
& K & Uniform & Non-uniform \\ \hline
Baseline  & - & 84.7 & 84.7 \\ \hline
\multirow{5}{*}{w M-HGC} & 3 & 86.4 ($\uparrow$ 1.7) & 86.0 ($\uparrow$ 1.3) \\
& 5 & 86.5 ($\uparrow$ 1.9) & 86.2 ($\uparrow$ 1.5) \\
& 7 & 86.3 ($\uparrow$ 1.6) & 86.5 ($\uparrow$ 1.8) \\
& 9 & 86.0 ($\uparrow$ 1.3) & \textbf{86.7 ($\uparrow$ 2.2)} \\ 
& 11 & 85.9 ($\uparrow$ 1.2) & 86.4 ($\uparrow$ 1.7) \\ \hline
% \multirow{5}{*}{Non-uniform}  & 3  & 86.0 ($\uparrow$ 1.3) \\
% & 5  & 86.2 ($\uparrow$ 1.5) \\
% & 7 & 86.5 ($\uparrow$ 1.8) \\
% & 9 & \textbf{86.7 ($\uparrow$ 2.2)} \\
% & 11 & 86.4 ($\uparrow$ 1.7) \\ \hline 
\end{tabular}
\caption{\textbf{Ablation of the hyper-parameter $K$ in A-NHG.} The Uniform represents K-uniform hyper-graph, in which one hyper-edge contains K joints. The non-uniform represents non-uniform hyper-graph, in which one joint is contained by $K$ hyper-edge.}
\label{comparison_A-NHG}
\end{table}

\begin{figure}
    \centering
    \includegraphics[width=\linewidth, trim=10mm 20mm 10mm 10mm]{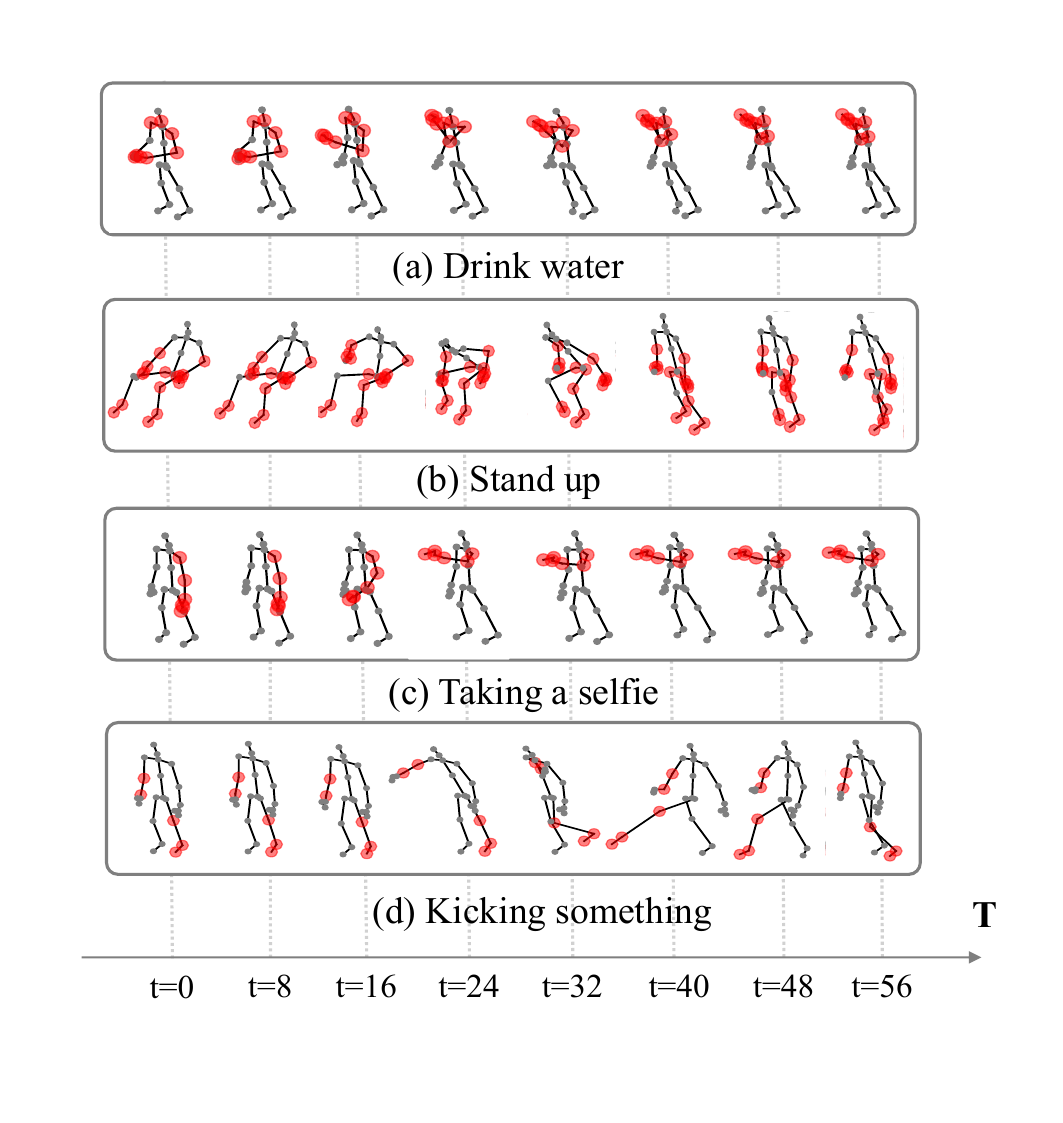}
    \caption{\textbf{Visualisation of hyper-graph in selected actions.} For ease of observation, the joints contained in the selected hyper-edge are highlighted in red.}
    \label{hyper-graph-visualise}
\end{figure}

\noindent \textbf{Hyper-graph Construction:}
The Non-uniform hyper-graphs with hyper-parameter $K$ represent the maximum number of hyper-edges contained in each joint.
They reflect the varying degrees of aggregation in the constructed topology. 
In general, it is common to adopt a uniform hyper-graph with hyper-parameter $K$.
Therefore, to explore how to set the hyper-parameter $K$ for the adaptive hyper-graph in M-HGC, 
we design the ablation experiments to investigate the impact of hyper-parameter $K$ on the model's recognition ability, where both uniform and non-uniform hyper-graphs are considered. 
% And we also compare the mode with uniform hyper-graph and Non-uniform hyper-graph.

The results are reported in Table \ref{comparison_A-NHG}. 
% When $K$ equals 1, each hyper-edge contains only one joint, which essentially eliminates any interaction between joints. 
% Therefore, we do not set K to 1.
% Observing the design of the single scale hyper-graph, it becomes clear that increasing $K$ does not always yield better performance. 
In both uniform and non-uniform hyper-graphs, a very large $K$ means that more joints are contained in each hyper-edge. 
Even though these "extra" joints may be optimised for minimal weight, they are likely to act as "noise",  disrupting the information interaction represented by the hyper-edge. 
As the $K$ decreases, it is difficult for the hyper-edge to represent complex joint combination patterns. 
% Joint with large differences in semantic information cannot interact sufficiently and information aggregation ability is lacking, which violates our original design intention. 
Therefore, setting $K = 5$ in uniform and $K = 9$ in Non-uniform achieves the best performance. The non-uniform hyper-graph of A-NHG outperforms the uniform hyper-graph, as the combination of skeleton joints in different actions is diverse. 

Furthermore, we visualise the hyper-graph inferred by Hyper-GCN on selected actions in Figure~\ref{hyper-graph-visualise}. 
The hyper-graph construction focuses on the joints that are most relevant to the corresponding action category. 
For “Kicking something” as part (d), the left leg joints and the right hand joints are modelled as a hyper-edge. Because in this action, the person's right hand is previously extended forward, and the left leg swings relatively to the right hand.
Similarly, in the action "Stand up" as shown in part (b), the hand and foot joints serve as fixed joints, while the remaining body joints move relative to them. 
In addition, "Taking a selfie" and "Drinking water" both focus on the main joints of movement with the above view as Figure~\ref{hyper-graph-visualise} (a), (c).
% For the complex and diverse characteristics of human action, multi-scale hyper-graphs in the spatial dimension can better capture features at multiple levels. 
% Therefore, we visualise the Top-1 accuracy of selected human action categories as shown in Figure. 
% This is especially advantageous for recognising actions with varying degrees of motion. 
% For instance, for actions like "Brushing teeth" or "Salutation", where only local joints are in motion, the smaller-scale hyper-graphs focus more on the local joints. 
% In contrast, for actions like "Stand up" and "Jump up", where the entire body is engaged, a larger-scale hyper-graph can capture information between distant joints. 
% This also inspired us to adopt the AHC-Module in different branches to complementarily learn hyper-graph topology at different scales.

\begin{table}
\centering
\begin{tabular}{cccc}
\hline
& & \multicolumn{2}{c}{NTU RGB+D 120 X-Sub (\%)} \\
& & w/o $\mathcal{L}_h$ & w/ $\mathcal{L}_h$  \\ \hline
Baseline & & 84.7 & 84.7 \\ \hline
\multirow{3}{*}{w/o M-HGC} & w/ 1 & 84.9 ($\uparrow 0.2$) & 84.9 ($\uparrow 0.2$) \\
& w/ 3 & 84.9 ($\uparrow 0.2$) & 85.2 ($\uparrow 0.5$) \\
& w/ 5 & 84.7 & 85.0 ($\uparrow 0.3$) \\ \hline
\multirow{3}{*}{w M-HGC} & w/ 1 & 86.7 ($\uparrow 2.0$) & 86.7 ($\uparrow 2.0$) \\
& w/ 3 & 86.6 ($\uparrow 1.8$) & \textbf{86.9} ($\uparrow 2.2$) \\
& w/ 5 & 86.6 ($\uparrow 1.8$) & 86.8 ($\uparrow 2.1$) \\ \hline

\end{tabular}
\caption{\textbf{Ablations on hyper joints.} w/ $N$ represents the number of hyper joints in each layer. w/ $\mathcal{L}_h$ represents the training with Divergence Loss. The hyper-parameter $K$ of  M-HGC is set to $9$.}
\label{virtual_comparison}
\end{table}

\begin{figure}
    \centering
    \includegraphics[width=\linewidth, trim=10mm 30mm 10mm 20mm]{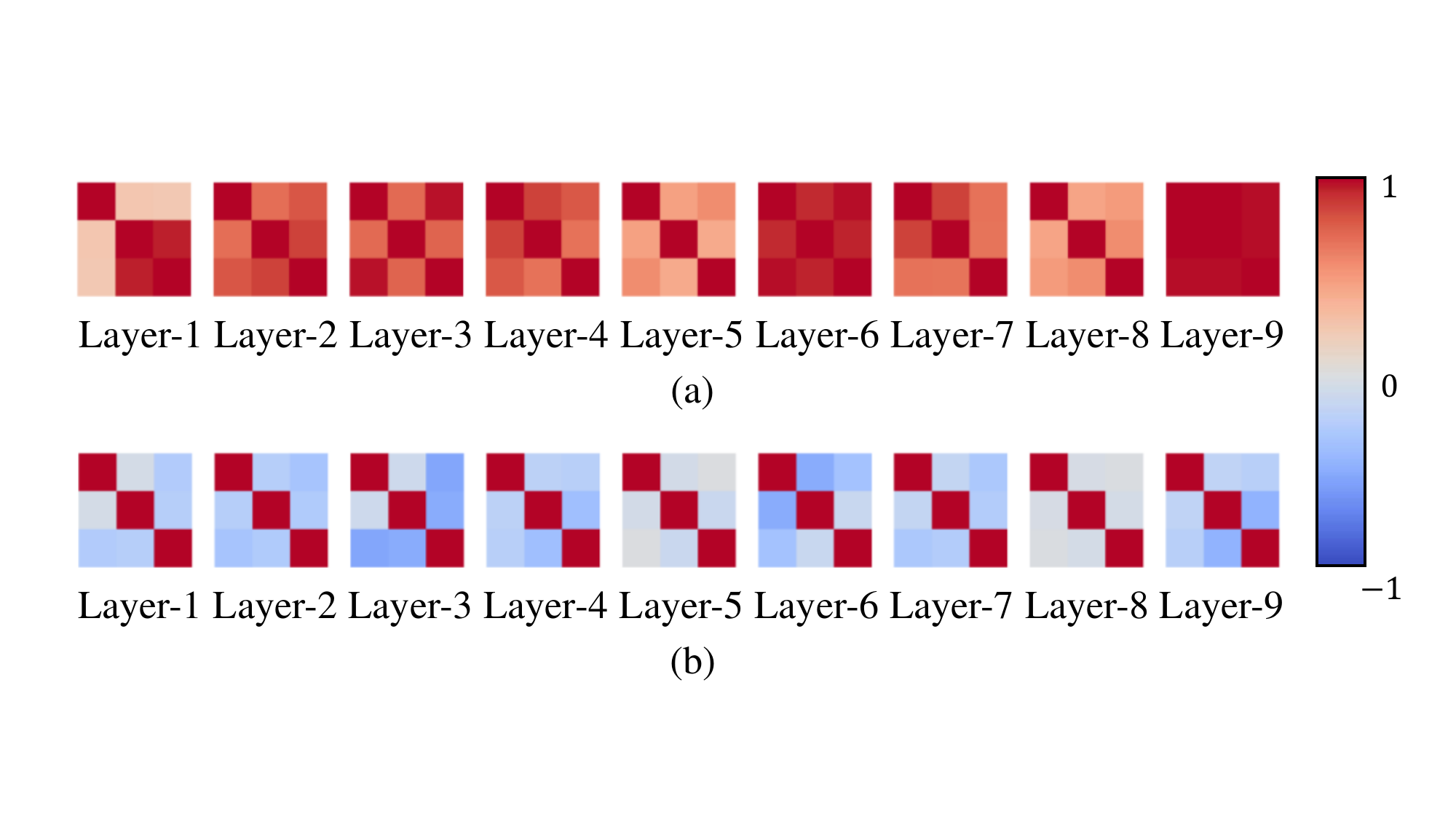}
    \caption{\textbf{Comparison of the cosine matrices of hyper-joints.} Part (a) and (b) represents w/o Divergence Loss and w/ Divergence Loss with 3 hyper-joints. It is calculated by Eqn.~\eqref{Eq-14}.}
    \label{matrix}
\end{figure}

\noindent \textbf{Virtual Connections:}
We conduct the ablation study to analyse the hyper joints and the Divergence Loss. 
The results are listed in Table~\ref{virtual_comparison}. 
By comparison, the performance is not monotonically increasing with the number of hyper joints.
The case of introducing only $3$ hyper joints achieves the best performance.
As our hyper joints are learned from a large amount of data, involving a large number of hyper joints can introduce redundant and ambiguous clues, degrading the performance. 

% In the baseline, we set the hyper-joints to be connected to the physical joints in a simple learnable topology. 
% Its performance with Divergence Loss is improved, similar to the setting with M-HGC. 
% We attribute to the fact that the topology in baseline is optimised based on the cross-entropy loss, which is not relevant to the difference between hyper-joints. Whereas with the addition of M-HGC, Divergence Loss guides the model to increase the difference between hyper-joints. And A-NHG constructs the hyper-graph based on the semantic information of them, which directly act on the hyper-graph topology constructed. 

% Besides, the involvement of hyper joints delivers consistent improvement, while introducing only a marginal increase in terms of model size and complexity.

% Building on this, we also experiment with the penalty weight of each stage for the hyper-joints at each stage. As the network depth increases, the penalty weight can further improve the model's performance. The reason for this improvement is that in deeper layers of the Hyper-GCN, the semantic information tends to converge, especially under the premise that the hyper-graph accelerates information interaction. As a result, the hyper-joints learned by the model tend to become more homogeneous. Therefore, to maintain the original intention of learning differentiated hyper-joints, it is important to place greater emphasis on the penalty term for the hyper-joints in the deeper layers of the network.

In addition, we visualise the cosine matrices of Hyper-GCN, as shown in Figure~\ref{matrix}. Clearly, in the absence of the Divergence Loss, the homogenisation of the learned hyper-joints is severe. 
This impedes the ability of hyper-joints to represent the generalised features of human actions.
% This indicates that the effect of introducing multiple hyper-joints into the hyper-graph convolution is nearly the same as introducing a single hyper-joint. 
The observation further validates the effectiveness of the Divergence Loss for hyper-joint optimisation, supporting our viewpoint that generalised features with a certain degree of differentiation are needed to participate in hyper-graph convolution.
% In addition, joints close to the border of the human, because the border joints are more varied across samples, it is not possible to modulate for the variability of the samples. 
% Whereas the connected joints close to the central joint are physically structurally closer, which cause that global information captured by the hyper joints is not significant enough. 

% We also visualize the topology between hyper joints and physical joints as shown in \textbf{Figure. ?}. Notice that the hyper joints are connected by XXX and XXX. As the network deeper, it gradually focuses on XXX and XXX. This is effective for a number of actions like XXX, and focusing on XXX enables the model to extract features in a more refine way.

% \begin{table}
%     \centering
%     \begin{tabular}{ccccc}
%     \hline
%     \multirow{2}{*}{Method} & \multirow{2}{*}{D} &\multirow{2}{*}{M-HGC} &\multirow{2}{*}{H} & NTU-RGB+D 120 \\
%     & & & & X-Sub \\ \hline
%     Baseline & & & & 84.7 \\ \hline
%     & $\checkmark$ & & & 84.9 ($\uparrow 0.2$) \\
%     & $\checkmark$ & & $\checkmark$ & 85.0 ($\uparrow 0.3$) \\
% Hyper-GCN & $\checkmark$ & $\checkmark$ & & 86.5 ($\uparrow 1.8$) \\
%     & $\checkmark$ & $\checkmark$ & $\checkmark$ & \textbf{86.9} ($\uparrow 2.2$) \\ \hline
%     \end{tabular}
%     \caption{\textbf{The ablation study of architecture settings.} D represents adding dense connection in backbone. M-HGC and H represent replacing with Mulit-head Hyper-graph Convolution and introducing hyper-joints.}
%     \label{model}
% \end{table}

\begin{figure}[t]
    \centering
    \includegraphics[width=\linewidth, trim=0mm 25mm 0mm 25mm]{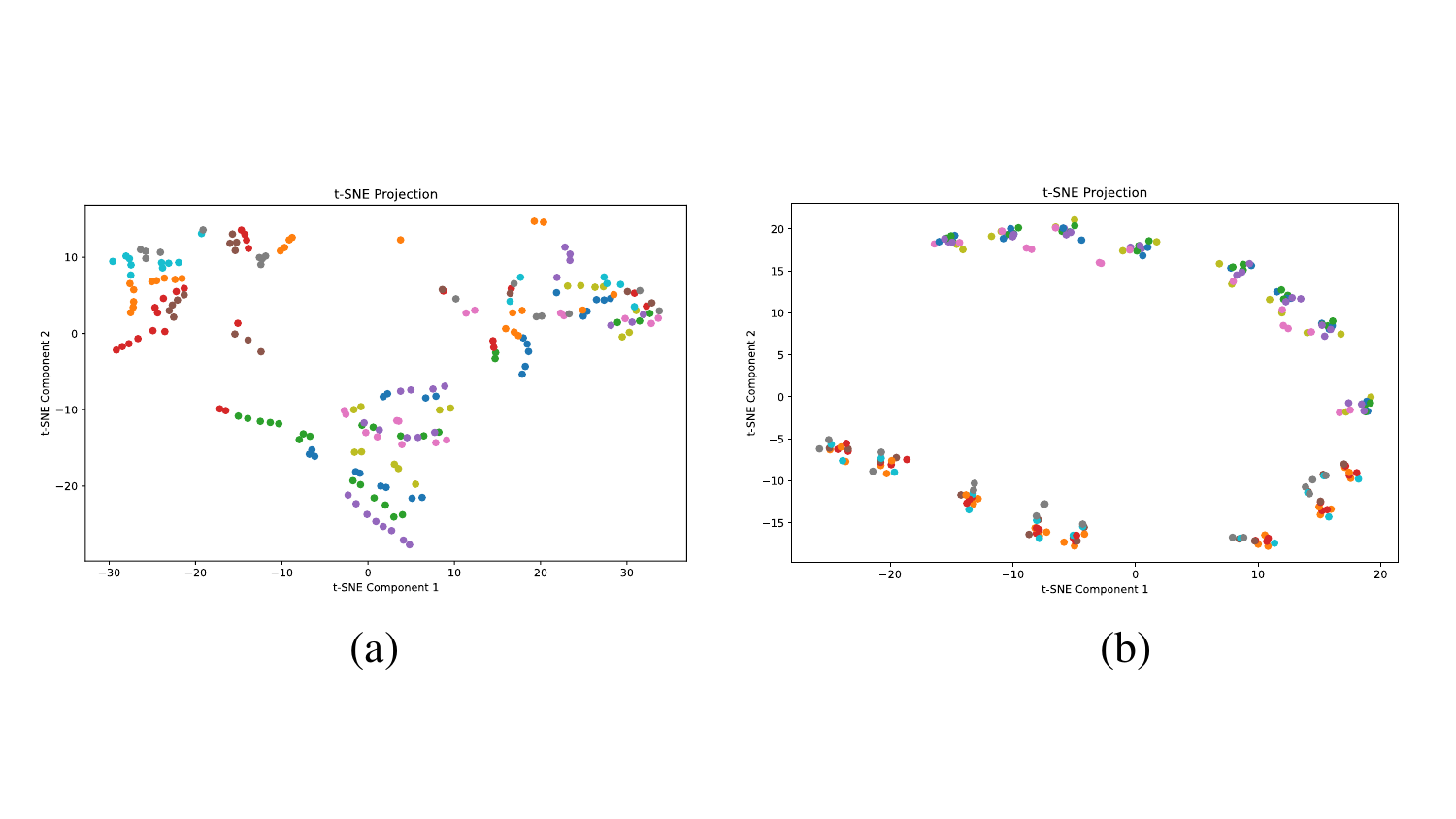}
    \caption{\textbf{t-SNE visualisation of Baseline and Hyper-GCN.} Different colours represent different joints. Part (a) and (b) reprsent the output features in Baseline and Hyper-GCN.}
    \label{t-SNE}
\end{figure}

\noindent \textbf{Effectiveness of Hyper-GCN:}
% We validate the effectiveness of performing dense connections, M-HGC and Virtual connections in the backbone.
% The experimental results are collected in Table~\ref{model}. 
% The dense connections in the experiments are only added within each stage. 
% It is observed that dense connections in shallow layers do not improve the performance as much as in deep layers. 
% As the layers go deeper, the number of channels gradually increases and the variance of information extracted from each layer becomes steeper. 
% This suggests that the deeper stages require dense connections to modulate the variance between features. 
% For example, the performance introducing dense connections at stage 1 and stage 2 is close to adding the dense connection only at stage 3. 
To further validate that Hyper-GCN increases the efficiency of information interaction, we perform t-SNE projections on the output features in last layer between Baseline and Hyper-GCN, as shown in Figure~\ref{t-SNE} for comparison. 
Notice that the semantic information represented by the last Hyper-GCN layer is very convergent. 
This represents that the semantic information of each joint is adequately conveyed in M-HGC.
The joints with similar semantics can prevent sacrificing the information after global average pooling.

\section{Conclusion}
In this paper, we propose an Adaptive Hyper-graph Convolutional Network for skeleton-based human action recognition. 
To exploit the implicit topology of multivariate synergy between joints, we introduce the Adaptive Non-uniform Hyper-graph, the Multi-head Hyper-graph Convolution and virtual connections. 
% We also introduce dense connections to fuse shallow and deep features. 
We carry out experiments on the dataset NTU-RGB+D 60 \& 120, NW-UCLA to validate the effectiveness of Hyper-GCN. 
The experimental analysis verifies that our design can improve the recognition performance. 
To the best of our knowledge, Hyper-GCN achieves the SOTA performance on 3 public datasets. 
The involvement of adaptive non-uniform hyper-graph modelling indeed extends existing GCN-based action recognition paradigms.

\noindent \textbf{Acknowledgment}
This work was supported in part by the National Natural Science Foundation of China (62020106012, 62106089, 62332008, 62336004), the Engineering and Physical Sciences Research Council (EPSRC), U.K. (Grants  EP/V002856/1, and EP/T022205/1), and the Fundamental Research Funds for the Central Universities (JUSRP202504007).

% \section{Final copy}

% You must include your signed IEEE copyright release form when you submit your finished paper.
% We MUST have this form before your paper can be published in the proceedings.

% Please direct any questions to the production editor in charge of these proceedings at the IEEE Computer Society Press:
% \url{https://www.computer.org/about/contact}.
{
    \small
    \bibliographystyle{ieeenat_fullname}
    \bibliography{main}
}

\end{document}